\documentclass{article} 
\usepackage{iclr2027_conference,times}
\iclrfinalcopy

\usepackage{amsmath,amsfonts,bm}

\def\eqref#1{equation~\ref{#1}}

\def\1{\bm{1}}

\DeclareMathAlphabet{\mathsfit}{\encodingdefault}{\sfdefault}{m}{sl}
\SetMathAlphabet{\mathsfit}{bold}{\encodingdefault}{\sfdefault}{bx}{n}

\usepackage{float}
\usepackage{hyperref}
\usepackage{url}
\usepackage{graphicx}
\usepackage{microtype}
\usepackage{booktabs}
\usepackage{amsmath,amssymb}
\usepackage{xcolor}
\usepackage{enumitem}
\usepackage{graphicx}
\usepackage{booktabs,tabularx,array}
\graphicspath{{figures/}}
\usepackage{textcomp}

\newcommand{\emailgroup}[2]{%
  {\normalfont\small\ttfamily
   \textbraceleft #1\textbraceright @#2}%
}

\title{RIFT: Residual-Informed Flow Transfer for One-Step Generative Video Compression}

\author{
Qizhe Wang\textsuperscript{1,*}
\quad
Jiaye Fu\textsuperscript{2,*}
\quad
Feng Ye\textsuperscript{3}
\quad
Haocheng Tang\textsuperscript{1}
\quad
Zetian Song\textsuperscript{1}
\\
\textbf{Chuanmin Jia}\textsuperscript{3}
\quad
\textbf{Siwei Ma}\textsuperscript{1}
\quad
\textbf{Jiaqi Zhang}\textsuperscript{1,\textdagger}
\\[4pt]
\textsuperscript{1}School of Computer Science, Peking University
\\
\textsuperscript{2}School of Electronic and Computer Engineering, Peking University
\\
\textsuperscript{3}Wangxuan Institute of Computer Technology, Peking University
\\[4pt]
\emailgroup{qzwang,jyfu,feng.ye,hctang,songzt}{stu.pku.edu.cn}
\\
\emailgroup{cmjia,swma,jqzhang}{pku.edu.cn}
}

\begin{document}

\maketitle

\footnotetext[1]{* Equal contribution.}
\footnotetext[2]{\textdagger Corresponding author.}
\footnotetext[3]{Preprint.}

\lhead{Preprint}

\begin{abstract}
Recent advances in video generation have enabled generative video compression to reconstruct details lost during coding using pretrained spatiotemporal priors.
However, a substantial gap remains between codec-induced degradation and the perturbation trajectories learned by flow-matching models.
In this paper, we identify a compression-dependent directional mismatch that degradation-strength-based timestep calibration alone cannot resolve.
Compressed latents remain strongly aligned across codec operating points, while their reconstruction residuals vary substantially in magnitude and direction.
To bridge this gap, we propose \textbf{RIFT}, a \textbf{R}esidual-\textbf{I}nformed \textbf{F}low \textbf{T}ransfer framework for one-step generative video compression.
At its core, RIFT introduces a \textbf{Residual Geometry Inducer} (RGI) that converts the pretrained flow prediction into a content-adaptive restoration update using decoder-visible codec cues.
RGI integrates flow-timestep conditioning with a learned full-band residual-strength estimate and content-conditioned direction and gain modulation, aligning the orientation and magnitude of the update with the compression-dependent residual geometry.
Extensive experiments on three public video compression benchmarks show that RIFT outperforms competing generative video compression methods. The code is publicly available on 
\href{https://github.com/qzwamg-web/RIFT-RESIDUAL-INFORMED-FLOW-TRANSFER-FOR-ONE-STEP-GENERATIVE-VIDEO-COMPRESSION}{RIFT}.

\end{abstract}

\section{Introduction}
\label{sec:intro}

Recent advances in video generation have produced pretrained models with powerful spatiotemporal priors \citep{wan2025wan}.
Generative video compression can exploit these priors to restore visually plausible detail that would be prohibitively expensive to transmit at low bitrates \citep{mao2025generative,ling2026free}.
Yet repeatedly evaluating a large generative model makes decoding costly.
One-step reconstruction offers a more practical path \citep{xue2025single,li2026yoda}, provided that the generative prior can be adapted to the errors of an actual video codec.

This adaptation is nontrivial because codec-induced artifacts differ from the perturbations encountered during generative pretraining \citep{relic2025bridging}.
Compression-aware conditioning and generative-state selection help bridge this gap \citep{mao2025generative,ling2026free}, but a scalar measure of degradation cannot, on its own, specify the vector-valued correction needed by a decoded latent.
In this paper, we revisit the geometry of codec residuals across compression operating points.
Although decoded latents at different operating points remain strongly aligned, their clean-minus-decoded residuals change appreciably in both magnitude and direction.
This \emph{compression-dependent directional mismatch} means that adjusting a flow timestep or scaling a shared update is insufficient to describe the desired restoration trajectory.

To address this mismatch, we propose \textbf{RIFT} (\textbf{R}esidual-\textbf{I}nformed \textbf{F}low \textbf{T}ransfer), a one-step generative video compression framework that adapts a pretrained video flow prior to the residuals of entropy-decoded latents.
Unlike timestep-only adaptation, RIFT does not assume that codec-dependent errors can be corrected by moving to a different point on an otherwise shared restoration trajectory.
RIFT therefore introduces the \textbf{Residual Geometry Inducer} (RGI), which turns a single backbone prediction into a content-adaptive correction whose direction and magnitude respond to local codec degradation.
RGI combines a learned full-band residual-strength estimate, a flow timestep selected by a fixed operating-point lookup table, and learned direction and gain modulation.
We design the lookup table to assign each codec operating point a reproducible backbone condition, while the learned components adjust the orientation and amplitude of its one-step update.
RGI additionally uses decoder-visible spatial codec cues to localize degradation without requiring an additional conditioning payload.

\begin{figure}[t!]
    \centering
    \includegraphics[width=\linewidth]
    {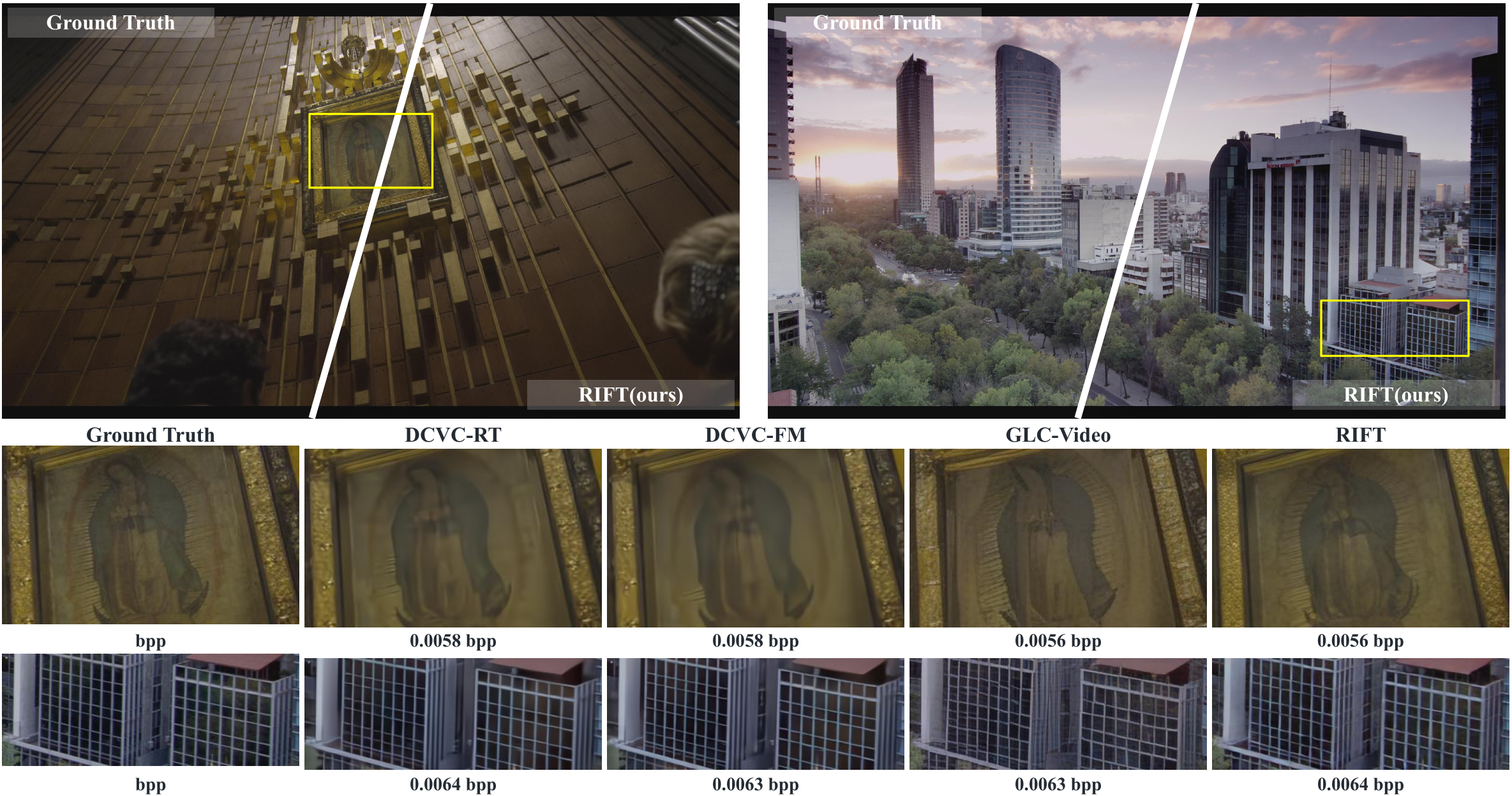}
    \caption{Qualitative comparison at ultra-low bitrates. Conventional and learned codecs produce blurred frames, while generative methods such as GLC-Video recover sharper textures but may introduce structural artifacts and inconsistent details. RIFT bridges the mismatch between compressed latents and the generative DiT through a lightweight codec--flow interface, restoring realistic textures while preserving structural fidelity in a single inference step. }
    \label{fig:teaser}
\end{figure}

We evaluate RIFT on HEVC-B, MCL-JCV, and UVG \citep{bossen2013common,wang2016mcl,mercat2020uvg} using actual entropy-coded bitstreams.
Across these three public video compression benchmarks, RIFT outperforms competing generative video compression methods, as shown in Fig.~\ref{fig:teaser}.

Our main contributions are summarized as follows.
\begin{itemize}
    \item
    We identify a compression-dependent directional mismatch in which decoded latents remain aligned across codec operating points while their reconstruction residuals differ substantially in direction and magnitude.

    \item
    We propose RIFT, a one-step generative video compression framework that directly restores entropy-decoded latents with a pretrained video flow prior adapted to compression-dependent residual geometry.

    \item
    We introduce RGI to bridge codec degradation and the pretrained flow trajectory through learned full-band residual-strength estimation, direction refinement, and gain modulation at a reproducible codec-conditioned flow timestep.

    \item
    Extensive experiments on HEVC-B, MCL-JCV, and UVG show that RIFT outperforms state-of-the-art generative video compression methods in perceptual quality, with RIFT's rates measured from actual entropy-coded bitstreams.
\end{itemize}

\section{Related Work}
\label{sec:related}

\subsection{Learned and Generative Video Compression}

Learned video compression has progressed from neural predictive coding to context-conditioned representation learning.
Early methods integrate motion compensation and residual coding into end-to-end optimization
\citep{lu2019dvc}.
Subsequent approaches exploit propagated features, multi-scale temporal contexts, and bidirectional references to improve conditional coding
\citep{li2021deep,sheng2021temporal,li2023neural,sheng2024spatial,sheng2024bidirectional}.
Complementary developments improve entropy estimation, rate adaptation, and practical efficiency through spatial-temporal probability modeling and efficient codec architectures
\citep{li2022hybrid,li2024neural,jia2025towards}.
These advances improve rate--distortion performance, while perceptual reconstruction at ultra-low bitrates motivates incorporating generative priors.

Generative compression uses adversarial objectives and learned generative representations to synthesize plausible details from compact bitstreams
\citep{mentzer2020high,yang2022perceptual,qi2025generative}.
Diffusion-based approaches further combine pretrained priors with temporal conditioning, semantic guidance, and efficient single-step decoding
\citep{ma2025diffusion,xue2025single,li2026yoda}.
Video-native priors extend this direction toward sequence-level refinement and trajectory-based compression
\citep{mao2025generative,ling2026free}, while quantization-aware diffusion formulations address discrepancies between codec degradation and generative pretraining
\citep{relic2025bridging}.
RIFT complements these approaches by explicitly modeling quantization-dependent residual geometry through decoupled timestep conditioning, direction prediction, and gain control.

\subsection{Video Generation Models}

Video generation has advanced through both discrete-token modeling and continuous diffusion.
Token-based approaches use masked or autoregressive transformers to model compressed video sequences conditioned on text and other modalities
\citep{villegas2022phenaki,kondratyuk2023videopoet}.
Diffusion-based methods extend image generators with temporal modeling and spatial-temporal upsampling to synthesize coherent, high-resolution videos
\citep{ho2022video,singer2022makeavideo,ho2022imagen}.
Latent diffusion further reduces computational demands and facilitates transferring pretrained image-generation capabilities to video
\citep{blattmann2023align,blattmann2023stable}.

Building on diffusion transformers
\citep{peebles2023scalable}, recent models combine spatiotemporal latent representations with joint or factorized attention to improve video generation
\citep{ma2024latte,yang2024cogvideox,zheng2024opensora}.
Large-scale training and flow-matching formulations further support powerful video foundation models with rich appearance and motion priors
\citep{lipman2022flow,kong2024hunyuanvideo,wan2025wan}.
These developments provide a strong foundation for generative reconstruction, but adapting pretrained models to compressed latents requires accounting for codec-specific degradation.
RIFT introduces a residual-informed interface around a pretrained video flow backbone, enabling codec-dependent correction within a single model evaluation.

\section{Compression-to-Flow Mismatch}
\label{sec:gap}

Let $\mathbf z$ denote the clean video latent and $\hat{\mathbf z}_q$ its
compressed observation at codec operating point $q$. Compression produces
rate-constrained, content-dependent residuals, whereas a pretrained flow model
follows a generative trajectory indexed by timestep $\tau$. Thus $q$ and $\tau$
describe different processes, and a flow timestep alone cannot determine the
direction and magnitude of $\mathbf z-\hat{\mathbf z}_q$.

\subsection{Compression Changes Residual Geometry}
\label{sec:gap:geometry}

The codec correction is the clean-minus-decoded residual
\begin{equation}
    \mathbf r_q=\mathbf z-\hat{\mathbf z}_q .
    \label{eq:codec_residual}
\end{equation}
If QP changed only degradation severity along a common trajectory, the
residuals would satisfy $\mathbf r_q\approx a_q\mathbf r_{\mathrm{ref}}$
for some scalar $a_q$. Measurements from actual codec encode--decode passes
contradict this approximation: decoded latents remain strongly aligned across
QPs, while their residuals vary substantially in both norm and direction.
Detailed measurements are provided in
Appendix~\ref{app:cross_qp_geometry}. Thus, a flow timestep can select a useful
operating point for the pretrained prior, but cannot specify the
content-dependent correction direction.

This does not rule out nonlinear flow trajectories or timestep-dependent
prediction directions. Rather, it shows that a shared-direction,
scalar-rescaling model is insufficient for codec residuals. RIFT therefore
factorizes the restoration process: the flow timestep selects a stable
operating point for the frozen prior, codec cues condition residual strength,
and learned corrections adjust the direction and magnitude of the resulting
update. These cues are used as conditioning evidence rather than direct
observations of the clean residual.

\subsection{Residual Strength Does Not Determine Update Magnitude}
\label{sec:gap:decoupling}

Residual strength alone cannot determine a restoration vector. For a predicted
update $\mathbf d$ and true residual $\mathbf r$, the gain minimizing endpoint
error is
\begin{equation}
    g^{*}=\frac{\langle\mathbf r,\mathbf d\rangle}
                 {\|\mathbf d\|_2^2}.
    \label{eq:optimal_gain}
\end{equation}
Hence, even at fixed residual strength, the optimal gain changes when the
predicted direction is imperfect. RIFT therefore separates residual strength,
direction correction, and gain modulation, while the flow timestep determines
the backbone operating point.

These roles form a decoder-compatible interface rather than quantities learned
in the same manner. RIFT estimates token-wise residual strength to condition
learned corrections, while the QP-to-flow-timestep mapping is specified by a
designed lookup table. Direction correction and gain modulation then refine the
backbone update, making each signal's role explicit without conflating residual
strength with a learned flow timestep.

\begin{figure}[t!]
    \centering
    \includegraphics[width=\linewidth]
    {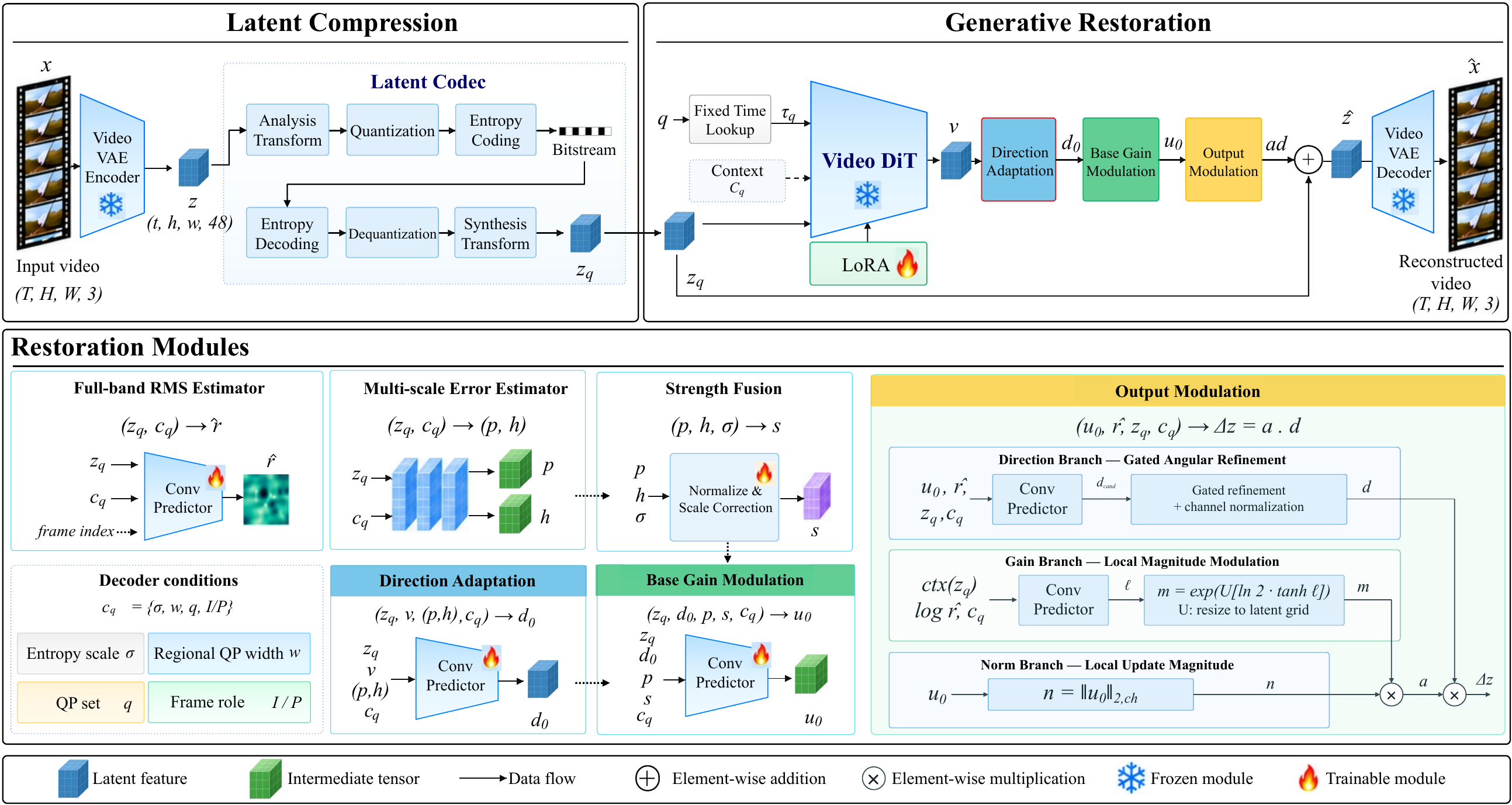}
    \caption{\textbf{Overview of RIFT.} The latent codec reconstructs the
compressed representation, and RGI adapts a single flow prediction using
decoder-visible conditions to restore it in one step.}
    \label{fig:pipeline}
\end{figure}

\section{Method}
\label{sec:method}

RIFT combines latent compression with one-step generative restoration.
A frozen video VAE encoder maps the input video $\mathbf x$ to
$\mathbf z=\mathcal E(\mathbf x)$.
The latent codec produces a bitstream and reconstructs
$\hat{\mathbf z}_q$ at operating point $q$.
The Residual Geometry Inducer (RGI) adapts a pretrained video flow
prediction to this decoded observation, producing a restored latent
$\tilde{\mathbf z}$.
The reconstructed video is
$\hat{\mathbf x}=\mathcal D(\tilde{\mathbf z})$, where
$\mathcal D$ is the frozen VAE decoder.
Figure~\ref{fig:pipeline} summarizes the framework.

\subsection{Residual Geometry Inducer}
\label{sec:interface}

\paragraph{Decoder conditions.}
RGI uses only the decoded latent and conditions available at the decoder:
\begin{equation}
    \mathbf c_q=\{\sigma,w,q,\rho\},
\end{equation}
where $\sigma$ is the entropy-model scale, $w$ is the regional
quantization width, and $\rho$ denotes the latent I/P role.
Normalized temporal position is derived internally.

The regional quantizer is a supporting component of the latent codec.
It predicts $w$ from entropy-model context shared by the encoder and decoder,
allowing the decoder to reproduce the same regional widths without
transmitting an additional dense map.
The regional QP width $w$ and entropy scale $\sigma$ provide complementary
information about local quantization and coding uncertainty.

\paragraph{Full-band RMS Estimator.}
The Full-band RMS Estimator predicts a spatially and temporally varying
residual-strength map:
\begin{equation}
\hat{\mathbf r}
=
\mathcal R_{\mathrm{RMS}}
\left(
\hat{\mathbf z}_q,\mathbf c_q,\eta
\right),
\end{equation}
where $\eta$ denotes the normalized frame index.
The prediction $\hat{\mathbf r}$ is directly supervised by the local RMS
magnitude of the clean-minus-decoded latent residual, providing an explicitly calibrated measure
of local codec degradation.

\paragraph{Multi-scale Error Estimator.}
In parallel, the Multi-scale Error Estimator extracts positive degradation
descriptors and shared contextual features:
\begin{equation}
(\mathbf p,\mathbf h)
=
\mathcal E_{\mathrm{err}}
\left(
\hat{\mathbf z}_q,\mathbf c_q
\right).
\end{equation}
Here, $\mathbf p$ contains positive error descriptors at multiple scales,
while $\mathbf h$ retains content-dependent contextual information.
Unlike the explicitly supervised RMS prediction, these features are learned
to support the subsequent restoration modules.

\paragraph{Strength Fusion.}
The Strength Fusion module combines the multi-scale descriptors with the
entropy-model scale:
\begin{equation}
\mathbf s
=
\mathcal S
\left(
\mathbf p,\mathbf h,\sigma
\right).
\end{equation}
Internally, this module normalizes the predicted descriptors and applies a
learned scale correction.
The resulting feature $\mathbf s$ represents codec-conditioned degradation
strength for Base Gain Modulation.
The two strength paths therefore serve complementary purposes:
$\hat{\mathbf r}$ provides explicitly supervised physical calibration,
whereas $(\mathbf p,\mathbf h,\mathbf s)$ provide contextual features for
restoration.

\paragraph{Fixed Time Lookup and one-step Video DiT prediction.}
The operating point selects a fixed Video DiT timestep through the
Fixed Time Lookup:
\begin{equation}
\tau_q=
\begin{cases}
520, & q=0,\\
640, & q=16,\\
800, & q=39,\\
960, & q=63.
\end{cases}
\end{equation}
Using the decoded latent, the selected timestep, and the embedded decoder
context, RGI queries the Video DiT once:
\begin{equation}
\mathbf v
=
\mathcal F_{\phi,\theta_{\mathrm{L}}}
\left(
\hat{\mathbf z}_q,\tau_q;\mathbf c_q
\right),
\end{equation}
where $\phi$ denotes the fixed foundation-model weights and
$\theta_{\mathrm{L}}$ denotes the trainable LoRA parameters.
Inference introduces no additional noise and requires neither iterative
denoising nor numerical flow integration.

\paragraph{Geometry adaptation and output modulation.}
The raw flow prediction $\mathbf v$ is first adapted to the
codec-dependent residual geometry. Decoder Adaptation predicts a correction
using the decoded latent, multi-scale error features, and decoder conditions,
while Base Gain Modulation converts the adapted direction into a restoration
proposal:
\begin{equation}
\begin{aligned}
\mathbf d_0
&=
\mathbf v+
\mathcal A
\left(
\hat{\mathbf z}_q,\mathbf v,\mathbf p,\mathbf h,\mathbf c_q
\right),\
\mathbf g_{\mathrm{base}}
&=
\mathcal G_{\mathrm{base}}
\left(
\hat{\mathbf z}_q,\mathbf d_0,\mathbf p,\mathbf s,\mathbf c_q
\right),\
\mathbf u_0
&=
\mathbf g_{\mathrm{base}}\odot\mathbf d_0.
\end{aligned}
\label{eq:rift_base_adaptation}
\end{equation}

Output Modulation then decomposes $\mathbf u_0$ into a normalized direction
and a local magnitude. The Norm Branch extracts
\begin{equation}
\mathbf n = \left\|\mathbf u_0\right\|_{2,\mathrm{ch}},
\qquad
\mathbf e_0 = \mathcal N(\mathbf u_0).
\label{eq:rift_local_norm}
\end{equation}
where $\mathcal N$ denotes channel-wise normalization at each spatiotemporal
latent position. The Direction Branch predicts a candidate direction and
applies a frame-role-dependent angular refinement:
\begin{equation}
\begin{aligned}
\mathbf d_{\mathrm{cand}}
&=
\mathcal G_{\mathrm{dir}}
\left(
\mathbf u_0,\hat{\mathbf r},\hat{\mathbf z}_q,\mathbf c_q
\right),\
\mathbf d
&=
\mathcal N
\left(
\mathbf e_0+
\beta_{\rho}
\left(
\mathbf d_{\mathrm{cand}}-\mathbf e_0
\right)
\right).
\end{aligned}
\label{eq:rift_direction_refinement}
\end{equation}

In parallel, the Gain Branch predicts a bounded local adjustment:
\begin{equation}
\boldsymbol{\ell}
= \mathcal G_{\mathrm{out}}\!\left(
\operatorname{ctx}(\hat{\mathbf z}_q),
\log\operatorname{sg}(\hat{\mathbf r}),
\mathbf c_q
\right),
\quad
\mathbf m
= \exp\!\left(
\mathcal U\!\left[
\log 2\cdot\tanh(\boldsymbol{\ell})
\right]
\right),
\quad
\mathbf a
= \mathbf n\odot\mathbf m.
\label{eq:rift_local_gain}
\end{equation}
where $\operatorname{sg}$ denotes stop-gradient and $\mathcal U$ resizes the
prediction to the latent grid. This parameterization constrains
$\mathbf m$ to $[1/2,2]$.

The final latent correction and restored latent are
\begin{equation}
\Delta\mathbf z=\mathbf a\odot\mathbf d,
\qquad
\tilde{\mathbf z}
=
\hat{\mathbf z}_q+\Delta\mathbf z.
\label{eq:rgi_update}
\end{equation}

This decomposition allows RGI to refine the restoration direction separately
from its local magnitude.

\subsection{Training Objectives}
\label{sec:training}

Training is divided into two curricula that separately optimize the latent
codec and the generative restoration pathway.

\paragraph{Codec curriculum.}
The codec curriculum optimizes the transmitted latent representation using
a combination of latent reconstruction and perceptual distortion:
\begin{equation}
\mathcal{D}_{\mathrm{codec}}
=
\mathcal{L}_{\mathrm{latent}}
+
\mathcal{L}_{\mathrm{LPIPS}},
\label{eq:codec_distortion}
\end{equation}
where $\mathcal{L}_{\mathrm{latent}}$ is the Smooth L1 distance between the
decoded latent $\hat{\mathbf z}_q$ and the clean latent $\mathbf z$.
The corresponding objective is
\begin{equation}
\mathcal{L}_{\mathrm{codec}}
=
R
+
\lambda_q\mathcal{D}_{\mathrm{codec}},
\label{eq:codec_objective}
\end{equation}
where $R$ is the differentiable entropy-rate estimate.
\paragraph{Generative restoration curriculum.}
The generative restoration curriculum optimizes the codec--flow pathway for
perceptual reconstruction. Let $\hat{\mathbf x}=\mathcal D(\tilde{\mathbf z})$
denote the restored video. Its distortion objective is
\begin{equation}
\mathcal{D}_{\mathrm{rest}}
=
0.2\,\mathcal{L}_{\mathrm{RGB}}
+
\mathcal{L}_{\mathrm{LPIPS}}
+
0.1\,\mathcal{L}_{\mathrm{DISTS}} + \mathcal{L}_{\mathrm{latent}},
\label{eq:restoration_distortion}
\end{equation}
where $\mathcal{L}_{\mathrm{RGB}}$ measures RGB L1 distortion, while
$\mathcal{L}_{\mathrm{LPIPS}}$ and $\mathcal{L}_{\mathrm{DISTS}}$ provide
complementary perceptual supervision.
The corresponding objective is
\begin{equation}
\mathcal{L}_{\mathrm{rest}}
= R+\lambda_q\mathcal{D}_{\mathrm{rest}}.
\label{eq:restoration_objective}
\end{equation}
Both curricula use the same rate--distortion weights:
$\lambda_q\in\{0.15,0.08,0.05,0.02\}$ for
$q\in\{0,16,39,63\}$, respectively.
Quantization is approximated by a straight-through estimator during
training, whereas all reported rates are measured from actual rANS-coded
bitstreams.

\section{Experiments}
\label{sec:experiments}

\subsection{Experimental Setup}
\label{sec:experimental_setup}

\paragraph{Datasets and protocol.}
We evaluate on HEVC Class B~\citep{bossen2013common},
MCL-JCV~\citep{wang2016mcl}, and UVG~\citep{mercat2020uvg},
comprising 42 sequences. For each sequence, we code the first 96
frames in four independently decoded groups of 25, 25, 25, and 21
frames. We evaluate QPs $\{0,16,39,63\}$ at both native
$1920\times1080$ resolution and $1280\times720$, obtained by resizing
the source RGB frames.

\paragraph{Implementation.}
We first train the conditional latent codec on windows from
Vimeo90K~\citep{xue2019video} and OpenVidHD~\citep{nan2025openvid}.
We then adapt the pretrained Wan video flow prior with rank-8 LoRA
and the decoder-visible RGI conditions described in
Section~\ref{sec:interface}.
Training is conducted on six NVIDIA H800 GPUs, with batch sizes of
$3$ and $1$ for $256\times256$ and $512\times512$ training crops,
respectively.
All reported RIFT reconstructions are obtained from independently
entropy-decoded rANS bitstreams rather than from a differentiable
quantization surrogate.

\paragraph{Metrics and comparisons.}
We compute rate from the actual coded payload and normalize it by
the total RGB pixel count of the encoded canvas, including codec
padding. We report RGB PSNR, LPIPS-VGG~\citep{zhang2018unreasonable},
DISTS~\citep{ding2020image}, and patch-based FID.
FID uses features extracted from $256\times256$ patches,
with statistics pooled across sequences within each dataset. Codec-only reconstruction from the same bitstream isolates the effect of RGI. GLC-Video~\citep{qi2025generative} and
GNVC-VD~\citep{mao2025generative} serve as the main external generative references. Cross-method rate--quality comparisons use matched frame ranges, resolutions, and metric definitions, and are
restricted to overlapping bitrate ranges. Same-QP comparisons are
used only for diagnosis.

\subsection{Main Results}
\begin{figure}[t!]
    \centering
    \includegraphics[width=\linewidth]
    {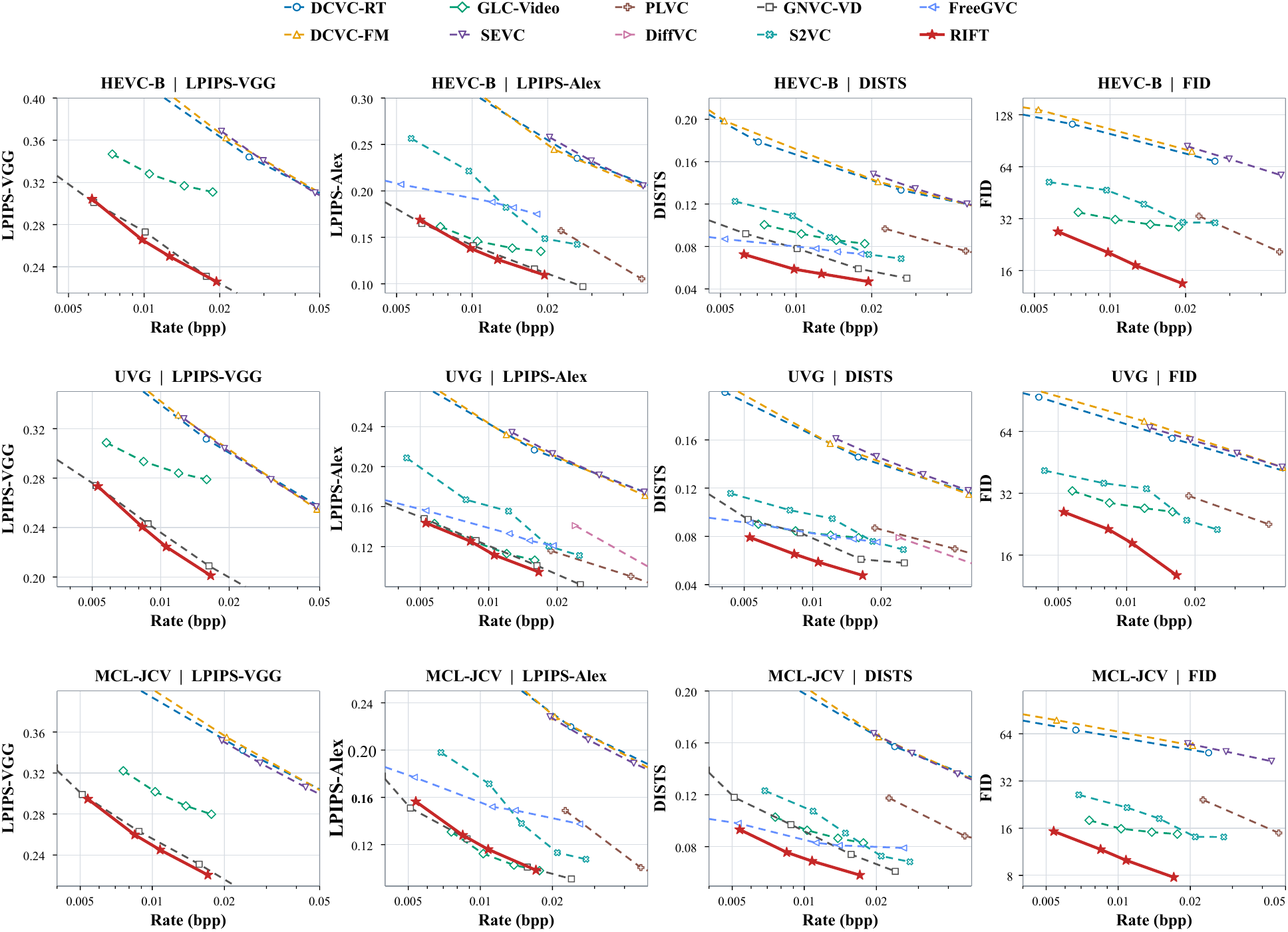}
    \caption{Rate--quality comparisons at 1080p on HEVC Class B,
    UVG, and MCL-JCV.}
    \label{fig:r-d}
\end{figure}

Fig.~\ref{fig:r-d} and Fig.~\ref{fig:rd_720} compare RIFT with
DCVC-RT~\citep{jia2025towards}, DCVC-FM~\citep{li2024neural}, SEVC~\citep{bian2025augmented}, PLVC~\citep{yang2022perceptual}, DiffVC~\citep{ma2025diffusion}, S2VC~\citep{xue2026single}, FreeGVC~\citep{ling2026free}, GLC-Video~\citep{qi2025generative}, and GNVC-VD~\citep{mao2025generative} at 1080p and 720p, respectively.
We locally reproduce and evaluate DCVC-RT, DCVC-FM, and GLC-Video, whereas
the results for the remaining baselines are taken from their corresponding
publications.
For GLC-Video, we follow its original inference protocol, which
conditionally codes each latent using the previously decoded latent as
temporal context and directly reconstructs each frame without iterative
diffusion refinement. The first 96 frames are evaluated with the
intra-period set to $-1$.
For GNVC-VD, we directly use the results reported in its original paper,
where each 96-frame sequence is divided into four groups of
$25/25/25/21$ frames and reconstructed using five flow-refinement steps.

Across the three datasets, RIFT achieves lower DISTS, LPIPS-VGG, and FID
than GLC-Video over their overlapping bitrate ranges.
Its advantage in DISTS remains consistent across both resolutions and
multiple operating points.

\subsection{Qualitative Results}
\begin{figure}[t!]
    \centering
    \includegraphics[width=\linewidth]
    {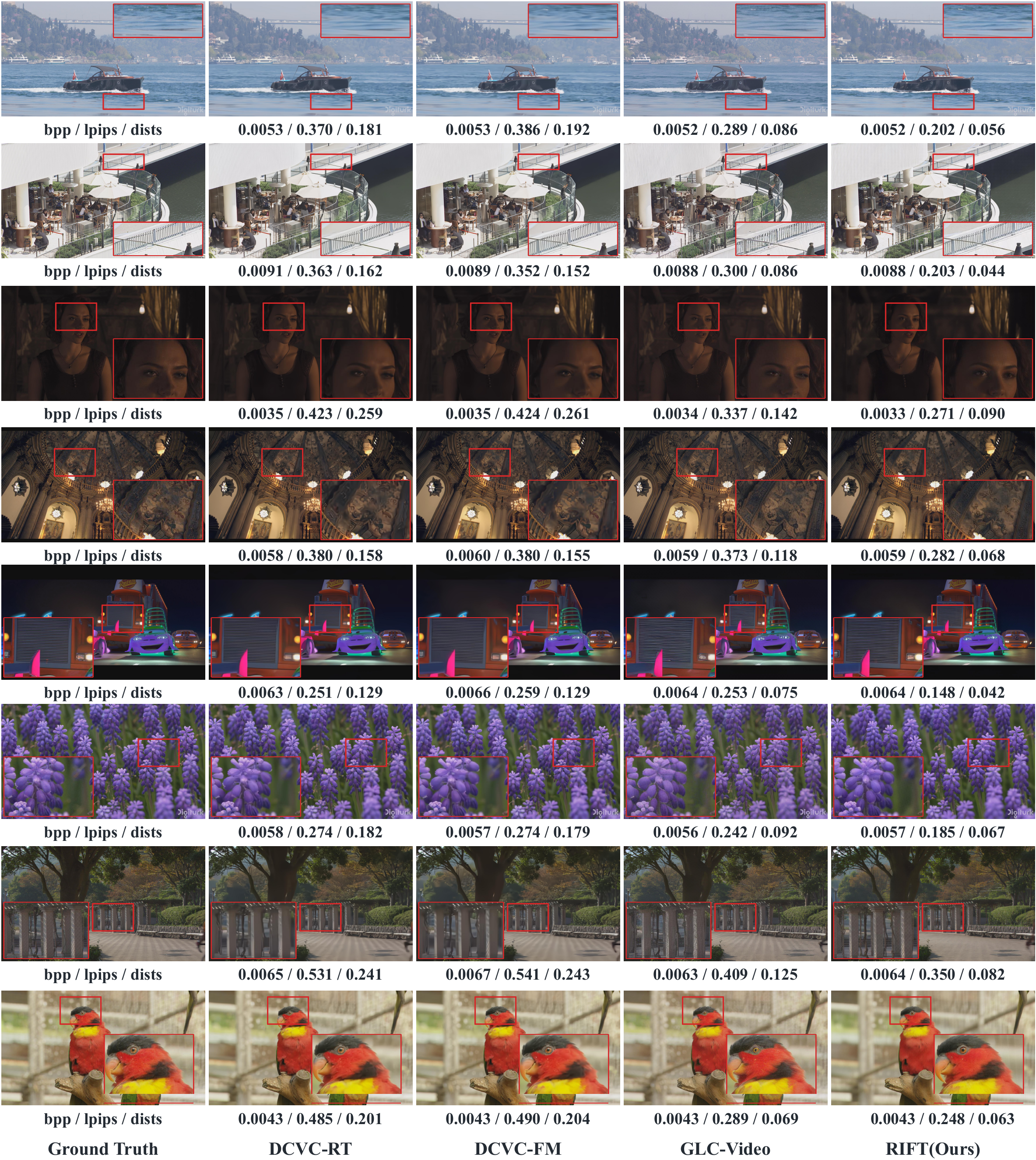}
    \caption{Qualitative comparison on eight sequences from HEVC
    Class B, UVG, and MCL-JCV. Red boxes indicate regions shown in
    enlarged insets.}
    \label{fig:sub}
\end{figure}

Fig.~\ref{fig:sub} compares RIFT with DCVC-RT, DCVC-FM, and
GLC-Video at closely matched bitrates. RIFT retains more distinct
surface patterns in the water scene and clearer thin boundaries
and repeated structures in the architectural scene. The facial
and vegetation examples likewise show recognizable contours and
fine appearance variations. Compared with DCVC-RT and DCVC-FM,
RIFT produces less smooth reconstructions; compared with
GLC-Video, its displayed details agree more closely with the
reference.

\paragraph{Ablation.}
We assess the dependence of the RIFT decoder on three restoration
components. BD-rate is measured relative
to RIFT over the common quality interval of dataset-averaged curves.

For the multi-scale error ablation, predicted errors are replaced by
the stored error reference and shared features are set to zero, while
scale correction and strength fusion remain active. For the direction
ablation, both the inherited direction residual and the subsequent
angular refinement are bypassed. For the gain ablation, learned base,
local, and global gain corrections are bypassed, while the fixed QP
gain table is retained and the final learned multiplier is set to one.
Each intervention is applied separately.

As shown in Table~\ref{tab:rift_module_ablation}, removing learned gain
modulation produces the largest penalty among the three module
interventions: $20.54\%$ BD-rate on LPIPS and $24.30\%$ on DISTS.
Bypassing multi-scale error estimation incurs penalties of $7.48\%$
and $9.76\%$, while bypassing direction refinement incurs $0.96\%$
and $3.76\%$, respectively. These results show that learned magnitude
modulation has the strongest effect among these inference-time
interventions.

We further evaluate a variant trained without the RGB-domain
reconstruction and perceptual losses, including RGB, LPIPS, and DISTS,
while retaining only latent-domain supervision. This variant yields
BD-rate increases of $50.95\%$ in LPIPS and $91.58\%$ in DISTS relative
to the full model, indicating substantial performance degradation.
These results show that latent-domain supervision alone is insufficient
to effectively guide generative restoration, whereas image-domain
reconstruction and perceptual objectives are essential for improving
the quality of the final decoded video.

\begin{figure*}[t]
\centering
\includegraphics[width=\textwidth,page=1]{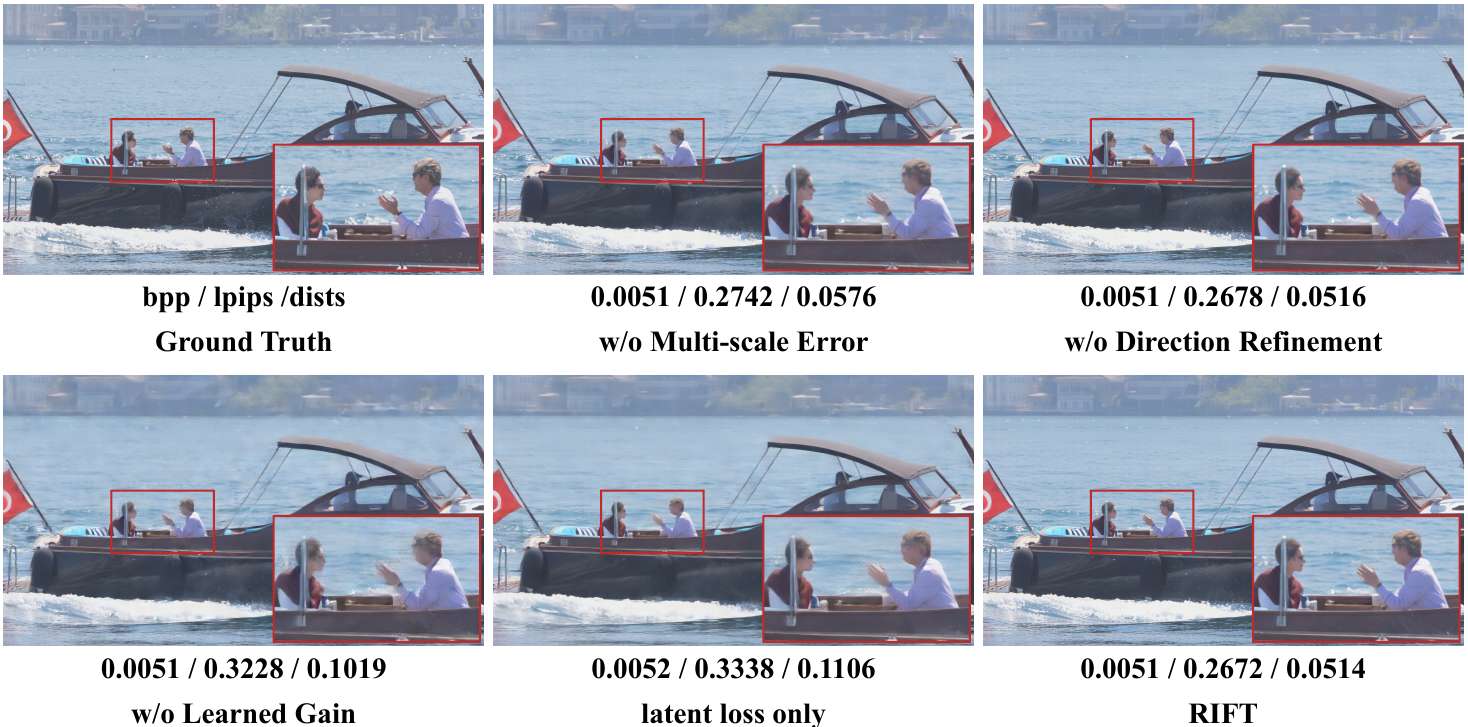}
\caption{Qualitative comparison of the ablation variants.
All variants are evaluated on the same frame and spatial crop.}
\label{fig:rift_module_ablation}
\end{figure*}

Figure~\ref{fig:rift_module_ablation} shows qualitative examples
with identical frames and crops across methods. The three module interventions use the same RIFT bitstream, whereas the latent-only variant is trained separately and its
bitrate is measured from its own entropy-coded bitstream. Additional codec-only comparisons are provided in Appendix~\ref{app:rift_codec_only}.

\begin{table}[t]
\centering
\small
\caption{Ablation on test sequences at 1080p. BD-rate is measured relative to full RIFT; positive values indicate increased rate at equal perceptual quality.}
\label{tab:rift_module_ablation}
\begin{tabular}{lrr}
\toprule
Configuration & BD-rate (LPIPS) & BD-rate (DISTS) \\
 & (\%) & (\%) \\
\midrule
w/o Multi-scale Error & +7.48 & +9.76 \\
w/o Direction Refinement & +0.96 & +3.76 \\
w/o Learned Gain & +20.54 & +24.30 \\
\bottomrule
\end{tabular}
\end{table}

\section{Conclusion}
\label{sec:conclusion}

We present \textbf{RIFT}, a Residual-Informed Flow Transfer framework for
one-step generative video compression. 
Our analysis shows that a scalar flow time alone is insufficient to describe
codec degradation, motivating the explicit separation of residual strength,
flow time, correction direction, and gain.
RIFT combines this factorized interface with rank-8 LoRA adaptation and
one-step reconstruction, enabling efficient use of a video foundation model
for compression restoration.
Experimental results demonstrate the effectiveness of modeling codec
residuals beyond scalar timestep alignment and show that RIFT outperforms
competing generative video compression methods.

\bibliography{refs}
\bibliographystyle{iclr2027_conference}

\clearpage
\appendix

\section{Additional Analysis of Codec Residuals}
\label{app:analysis}

We analyze decoded-latent residuals from actual codec encode--decode
passes. These diagnostics support
Section~\ref{sec:gap} and do not measure RGB reconstruction quality.

\subsection{Decoded-Latent Residual Distribution}
\label{app:residual_dist}

\paragraph{Protocol.}
For a clean latent $\mathbf z$ and its decoded counterpart
$\hat{\mathbf z}_q$ at QP $q$, we analyze the codec residual
$\mathbf r_q$ defined in Eq.~(\ref{eq:codec_residual}).
At each input resolution ($256\times256$ and $512\times512$), we use
32 fixed latent windows and QPs $\{0,16,39,63\}$, separating I- and
P-frame positions. Each QP/frame-role group contributes up to
$300{,}000$ scalar samples; channel-wise analysis uses up to
$20{,}000$ samples per channel across 48 channels.

We compare each empirical distribution with a maximum-likelihood
Gaussian fit and a uniform reference whose bounds are derived from
the codec quantization scale. We quantify their discrepancy using
the Kolmogorov--Smirnov (KS) distance,
\begin{equation}
    D_n = \sup_x \left|F_n(x)-F_{\mathrm{ref}}(x)\right|,
    \label{eq:app_ks}
\end{equation}
where $F_n$ and $F_{\mathrm{ref}}$ denote the empirical and reference
CDFs. We use the KS statistic only as a descriptive measure of distributional
mismatch, rather than as a hypothesis test. We report it for both the raw
residuals and the residuals normalized by decoder-visible quantization
widths.

\begin{figure}[t]
    \centering
    \includegraphics[width=\linewidth]{
        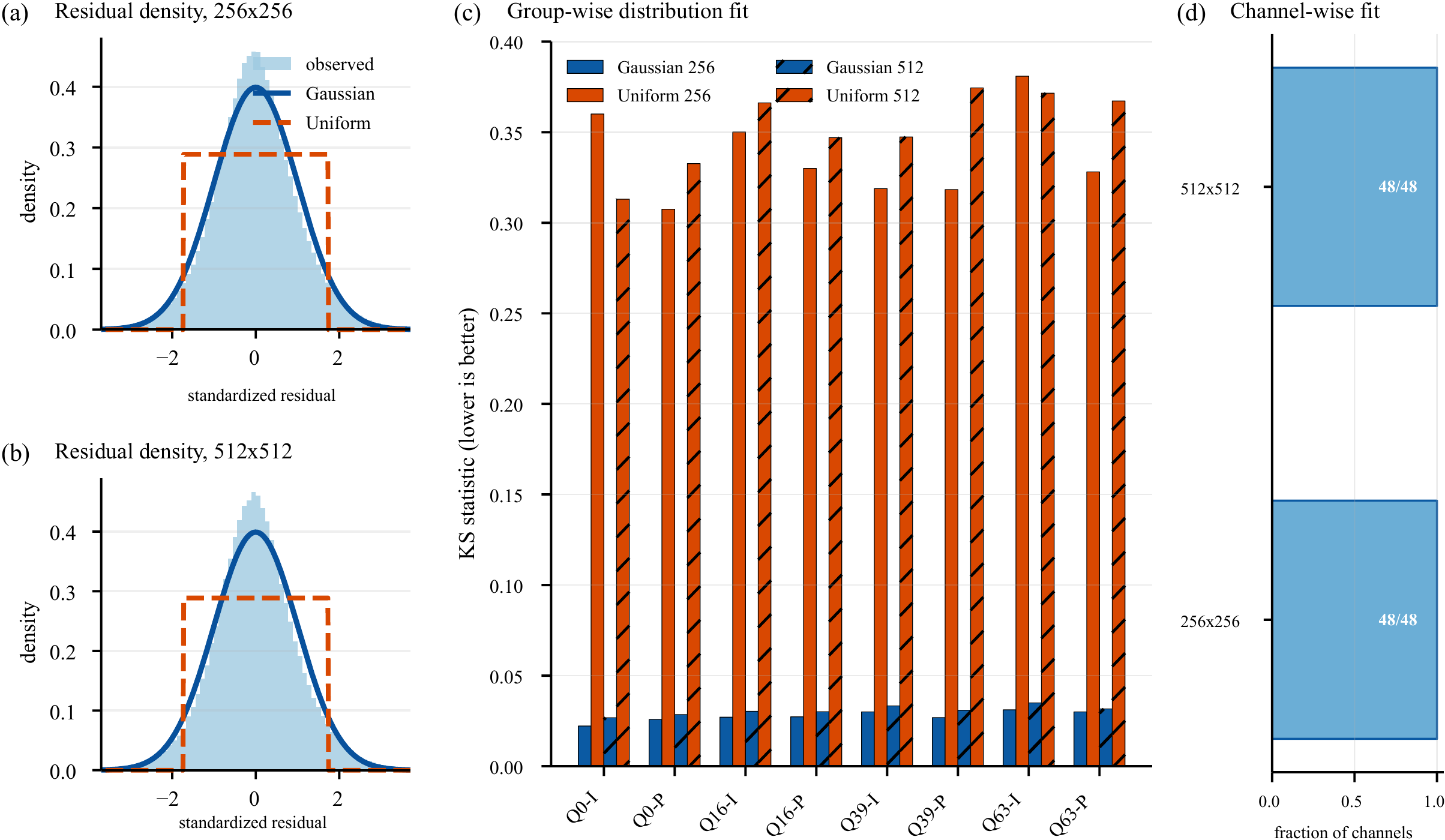
    }
    \caption{Decoded-latent residual distributions. (a--b) Standardized empirical densities with fitted Gaussian and scale-derived uniform references at $256\times256$ and $512\times512$. (c) KS distances by QP and frame role. (d) Fraction of channels favoring the Gaussian.}
    \label{fig:residual_distribution}
\end{figure}

\paragraph{Results.}
At both resolutions, the fitted Gaussian has a lower KS distance
than the scale-derived uniform reference in all eight QP/frame-role
groups, with and without width normalization
(Table~\ref{tab:residual_distribution_ks}). The channel-wise
comparison also favors the Gaussian across all 48 channels at each
resolution (Figure~\ref{fig:residual_distribution}).

\paragraph{Limits.}
Residuals approximate a Gaussian but deviate measurably:
skewness ranges from $0.09$ to $0.24$, and excess kurtosis from
$0.64$ to $1.32$. These are residuals after complete codec decoding,
not isolated scalar quantization errors. Marginal fit does not
establish independence, spatial whiteness, isotropy, or restoration
direction. It can inform degradation-strength estimation, while
cross-QP directions require separate analysis.

\subsection{Cross-QP Geometry of Codec Residuals}
\label{app:cross_qp_geometry}

\paragraph{Protocol.}
At each resolution, we encode and decode the same 512 clean latent
windows independently at all four QPs. We compare the compressed
latents $\hat{\mathbf{z}}_q$ and residuals $\mathbf{r}_q$ defined in Eq.~(\ref{eq:codec_residual}) as window vectors and corresponding
token vectors. For nonzero vectors,
\begin{equation}
    c(\mathbf{a},\mathbf{b})
    =
    \frac{\langle\mathbf{a},\mathbf{b}\rangle}
    {\|\mathbf{a}\|_2\|\mathbf{b}\|_2},
    \label{eq:app_cosine}
\end{equation}
and the corresponding angle is
\begin{equation}
    \theta(\mathbf{a},\mathbf{b})
    =
    \arccos c(\mathbf{a},\mathbf{b}).
    \label{eq:app_angle}
\end{equation}
Latent cosine measures representation alignment; residual cosine
measures alignment of corrections toward the common clean target.

\begin{figure}[t]
    \centering
    \includegraphics[width=\linewidth]{
        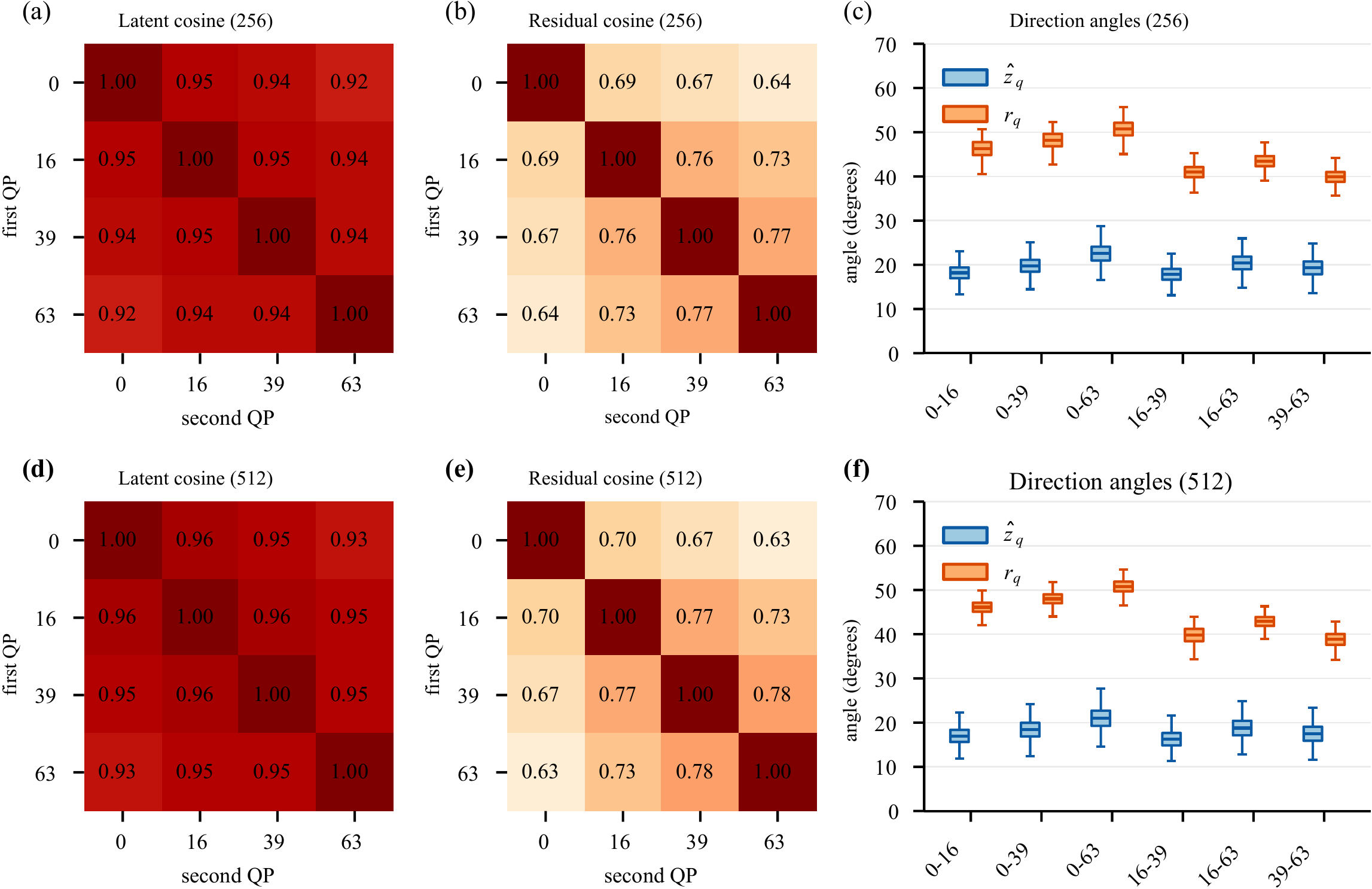
    }
    \caption{Cross-QP geometry at $256\times256$ (a--c) and $512\times512$ (d--f). Columns show compressed-latent cosine similarity, residual cosine similarity, and pairwise direction angles. Each resolution uses 512 matched latent windows.}
    \label{fig:cross_qp_geometry}
\end{figure}

\paragraph{Results.}
Mean cross-QP cosine similarity exceeds $0.94$ for compressed
latents but is about $0.71$ for residuals, corresponding to mean
residual angles near $44^\circ$
(Table~\ref{tab:cross_qp_geometry}). For the $q=0$ and $q=63$
pair, latent cosine values are $0.9236$ and $0.9331$ at the two
resolutions, whereas residual cosine is about $0.635$ and the
angle is near $50.5^\circ$ (Figure~\ref{fig:cross_qp_geometry}).
Shared latent content therefore does not imply aligned corrections.

\begin{table}[t]
    \centering
    \small
    \caption{Mean KS distances over eight QP/frame-role groups. Gaussian parameters are fitted to the residuals; uniform bounds follow the quantization scale. ``Wins'' denotes the number of groups for which the Gaussian has a lower KS distance.}
    \label{tab:residual_distribution_ks}
    \begin{tabular}{llccc}
        \toprule
        Resolution
        & Residual
        & Gaussian KS
        & Uniform KS
        & Wins \\
        \midrule
        $256\times256$
        & Raw
        & 0.0273
        & 0.3465
        & 8/8 \\
        $256\times256$
        & Width-normalized
        & 0.0272
        & 0.3481
        & 8/8 \\
        $512\times512$
        & Raw
        & 0.0306
        & 0.3431
        & 8/8 \\
        $512\times512$
        & Width-normalized
        & 0.0309
        & 0.3426
        & 8/8 \\
        \bottomrule
    \end{tabular}
\end{table}

\begin{table}[t]
    \centering
    \small
    \caption{Cross-QP cosine similarity and mean angle. Global and token cosine values average off-diagonal pairs at the window and corresponding-token levels, respectively; angles are measured at the window level.}
    \label{tab:cross_qp_geometry}
    \begin{tabular}{llccc}
        \toprule
        Resolution
        & Representation
        & Global cosine
        & Mean angle
        & Token cosine \\
        \midrule
        $256\times256$
        & Compressed latent
        & 0.9412
        & $19.57^\circ$
        & 0.9340 \\
        $256\times256$
        & Residual
        & 0.7080
        & $44.76^\circ$
        & 0.7056 \\
        \midrule
        $512\times512$
        & Compressed latent
        & 0.9496
        & $18.08^\circ$
        & 0.9419 \\
        $512\times512$
        & Residual
        & 0.7140
        & $44.25^\circ$
        & 0.7195 \\
        \bottomrule
    \end{tabular}
\end{table}

\section{Training Curriculum}
\label{app:training_curriculum}

The training curriculum comprises codec representation learning
and generative restoration adaptation.
It progresses from temporal compression pretraining to interface
warm-up, LoRA adaptation, joint optimization, and high-resolution
refinement.
Tables~\ref{tab:codec_curriculum}
and~\ref{tab:restoration_curriculum} specify the training budgets
and nominal learning rates.

\paragraph{Codec curriculum.}
The codec first learns temporal compression from seven-frame
RGB sequences in Vimeo-90K for 30K iterations.
It subsequently undergoes multi-rate adaptation on OpenVidHD
at $256\times256$ resolution for 20K iterations.
OpenVidHD training uses windows containing seven temporal
positions in the VAE latent representation, denoted by
$T_{\mathrm{latent}}=7$.

\begin{table*}[t]
    \centering
    \small
    \setlength{\tabcolsep}{5pt}
    \renewcommand{\arraystretch}{1.15}
    \caption{Codec learning curriculum.
    Learning rates denote nominal values before scheduling.}
    \label{tab:codec_curriculum}
    \begin{tabularx}{\textwidth}{
        @{}p{0.22\textwidth}
        p{0.25\textwidth}
        r
        >{\raggedright\arraybackslash}X@{}}
        \toprule
        Stage & Training data & Iterations
        & Trainable modules and learning rates \\
        \midrule
        Compression pretraining
        & Vimeo-90K, seven RGB frames
        & 30K
        & Latent codec: $10^{-4}$ \\
        \addlinespace
        Multi-rate adaptation
        & OpenVidHD, $256\times256$,
          $T_{\mathrm{latent}}=7$
        & 20K
        & Codec: $3\times10^{-6}$;
          regional quantizer: $3\times10^{-5}$ \\
        \bottomrule
    \end{tabularx}
\end{table*}

\paragraph{Generative restoration curriculum.}
We first calibrate the operating-point-to-time mapping using
a calibration subset of the training data.
The context injection modules and connection heads then undergo
a short warm-up with the codec and DiT backbone frozen.
LoRA adaptation subsequently enables the generative prior
to learn compression-residual restoration while the codec
provides a stable input distribution.

The joint optimization stage updates the codec together with
the restoration modules, allowing the compressed representation
and generative decoder to adapt to each other.
Training finally progresses to $512\times512$ windows with
lower learning rates to refine high-resolution reconstruction.
The VAE and base DiT weights remain frozen throughout;
DiT adaptation uses LoRA.

\begin{table*}[t]
    \centering
    \small
    \setlength{\tabcolsep}{5pt}
    \renewcommand{\arraystretch}{1.15}
    \caption{Generative restoration curriculum.
    Time calibration involves no gradient updates.
    Context denotes context injection and decoder adaptation
    modules; heads denote restoration connection heads.}
    \label{tab:restoration_curriculum}
    \begin{tabularx}{\textwidth}{
        @{}p{0.19\textwidth}
        p{0.24\textwidth}
        r
        >{\raggedright\arraybackslash}X@{}}
        \toprule
        Stage & Training data & Iterations
        & Trainable modules and learning rates \\
        \midrule
        Time calibration
        & OpenVidHD calibration subset,
          $256\times256$, $T_{\mathrm{latent}}=7$
        & ---
        & No parameter updates \\
        \addlinespace
        Interface warm-up
        & OpenVidHD, $256\times256$,
          $T_{\mathrm{latent}}=7$
        & 1K
        & Heads and context: $10^{-4}$ \\
        \addlinespace
        LoRA adaptation
        & OpenVidHD, $256\times256$,
          $T_{\mathrm{latent}}=7$
        & 30K
        & LoRA, heads, and context: $10^{-4}$ \\
        \addlinespace
        Joint optimization
        & OpenVidHD, $256\times256$,
          $T_{\mathrm{latent}}=7$
        & 30K
        & Codec: $2\times10^{-6}$;
          quantizer: $10^{-5}$;
          strength/gain heads: $3\times10^{-5}$;
          direction head: $10^{-5}$;
          inherited heads/context: $5\times10^{-6}$;
          LoRA: $10^{-5}$ \\
        \addlinespace
        High-resolution refinement
        & OpenVidHD, $512\times512$,
          $T_{\mathrm{latent}}=7$
        & 8K
        & Codec: $10^{-6}$;
          quantizer: $5\times10^{-6}$;
          strength/gain heads: $2\times10^{-5}$;
          direction head: $10^{-5}$;
          inherited heads/context: $3\times10^{-6}$;
          LoRA: $5\times10^{-6}$ \\
        \bottomrule
    \end{tabularx}
\end{table*}

\paragraph{Optimization and supervision.}
Optimization uses AdamW with linear warm-up followed by
cosine learning-rate decay.
The strength estimator receives local full-band residual
RMS supervision, while the direction and gain branches
control the orientation and magnitude of the restoration
update.
Restoration training combines latent-space, RGB, and perceptual
objectives.
Joint optimization additionally includes the codec
rate--distortion objective.
At inference, the fixed operating-point-to-time lookup
conditions a single DiT evaluation per latent window.

\section{Additional Quantitative Results}
\label{app:additional_results}

\subsection{Results at 720p}
\label{app:results_720p}

\begin{figure*}[t]
\centering
\includegraphics[width=\textwidth]
{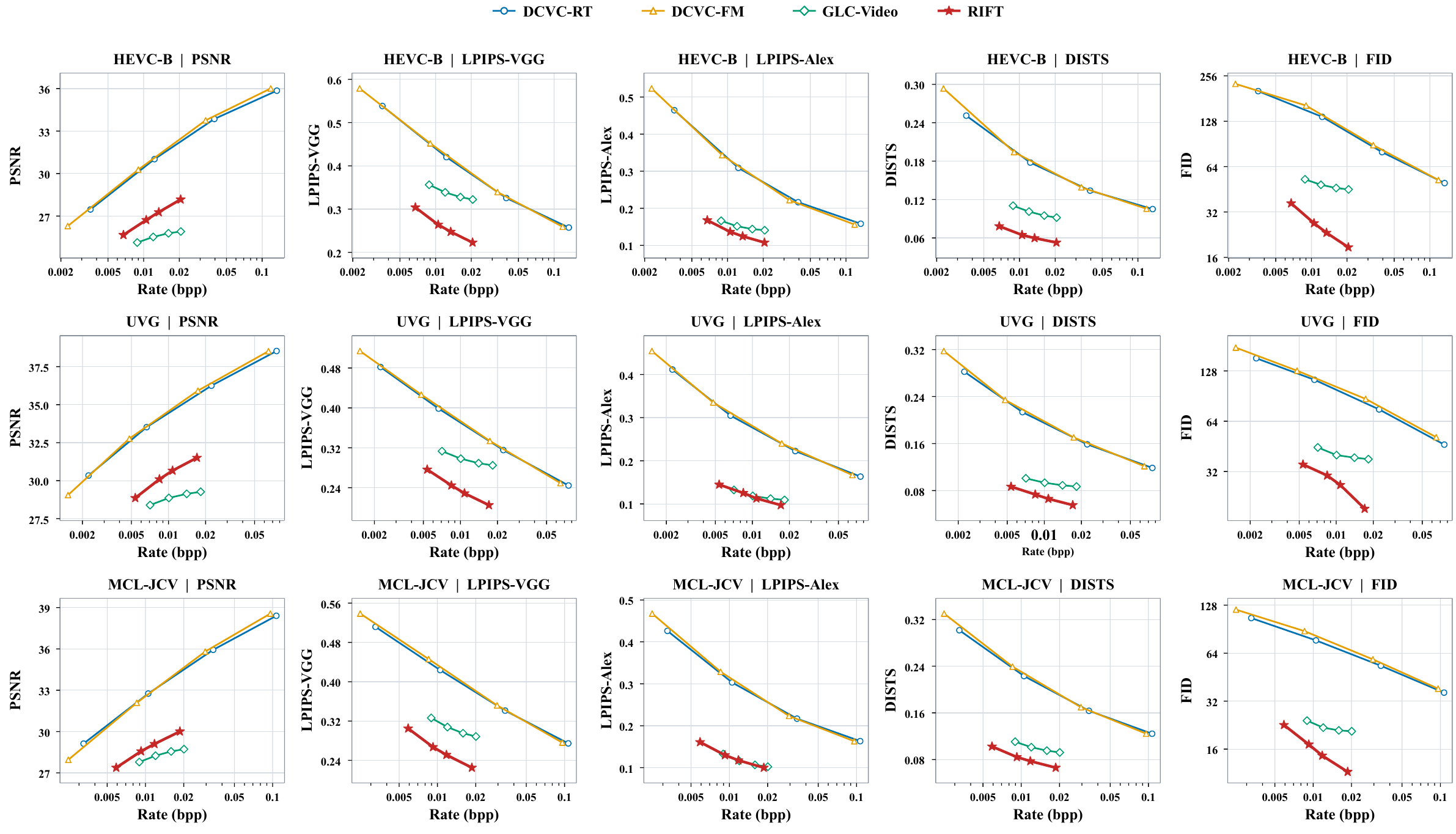}
\caption{Rate--quality comparisons at 720p on HEVC-B, UVG, and
MCL-JCV. DCVC-RT, DCVC-FM, and GLC-Video are locally reproduced
and evaluated under the same test setting. The reported metrics
are PSNR, LPIPS-VGG, LPIPS-Alex, DISTS, and FID. Higher PSNR and
lower values of the perceptual metrics indicate better
performance.}
\label{fig:rd_720}
\end{figure*}

Figure~\ref{fig:rd_720} presents the rate--quality results at 720p
on HEVC-B, UVG, and MCL-JCV. For this comparison, we locally
reproduce DCVC-RT, DCVC-FM, and GLC-Video and evaluate them together
with RIFT under the same test setting. The two distortion-oriented
codecs, DCVC-RT and DCVC-FM, achieve substantially higher PSNR, but
their perceptual metrics deteriorate rapidly toward the low-rate
regime. This behavior reflects the expected distortion--perception
trade-off: optimizing pixel-wise fidelity tends to produce smooth
reconstructions when the bitrate becomes insufficient to preserve
fine visual details.

Compared with GLC-Video over their overlapping bitrate ranges, RIFT
consistently achieves lower LPIPS-VGG, LPIPS-Alex, DISTS, and FID
across the three datasets. The improvements are particularly clear
for DISTS and FID, indicating better preservation of perceptually
important structures and more realistic reconstructed-frame
statistics. RIFT also provides higher PSNR than GLC-Video throughout
the common rate range. Although its PSNR remains below that of the
distortion-oriented codecs, the simultaneous improvement over
GLC-Video in both pixel fidelity and perceptual quality suggests that
the gain does not arise merely from trading distortion for stronger
generative synthesis. Instead, the residual-informed adaptation
allows the pretrained flow prior to produce details that remain more
consistent with the decoded video content.

\subsection{PSNR Results at 1080p}
\label{app:psnr_1080p}

\begin{figure*}[t]
\centering
\includegraphics[width=\textwidth]
{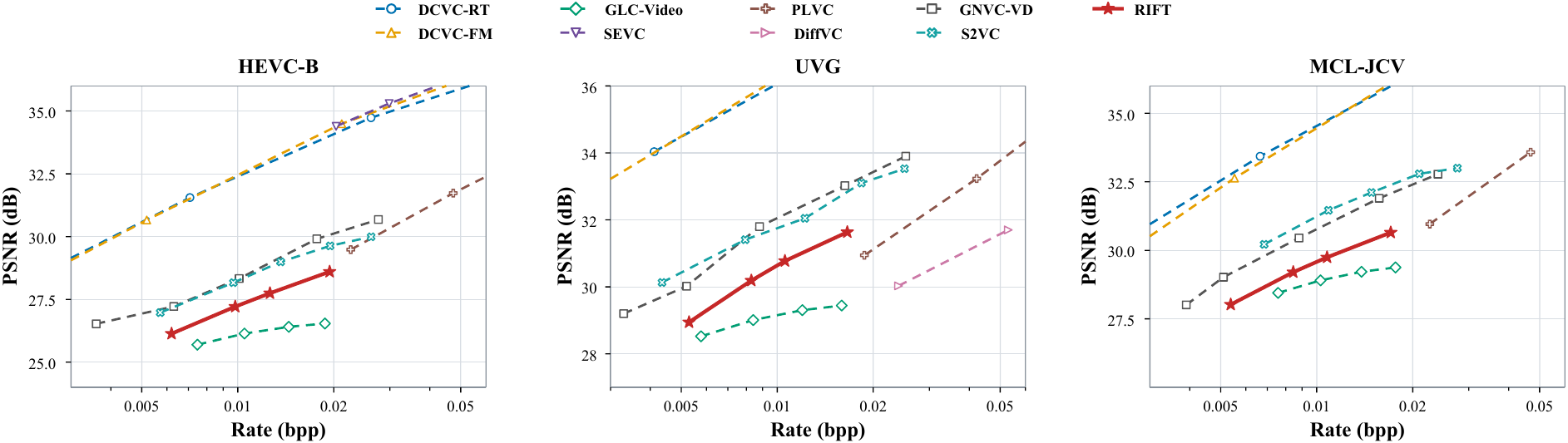}
\caption{PSNR--bitrate comparisons at 1080p on HEVC-B, UVG, and
MCL-JCV. RIFT improves upon GLC-Video over their overlapping
bitrate ranges, although its gains are more pronounced in
perceptual quality than in pixel-level fidelity. Higher PSNR is
better.}
\label{fig:psnr_1080}
\end{figure*}

Figure~\ref{fig:psnr_1080} further reports the PSNR--bitrate results
at 1080p. RIFT consistently improves upon GLC-Video over their common
bitrate ranges on HEVC-B, UVG, and MCL-JCV, showing that the proposed
restoration does not obtain its perceptual advantage at the cost of
additional pixel-level distortion relative to this generative
baseline.

The PSNR improvement is nevertheless more moderate than the gains
observed in perceptual metrics and qualitative comparisons. This is
consistent with the design of RIFT: its lightweight interface
modulates the direction and magnitude of the update predicted by the
pretrained Video DiT, rather than introducing a large auxiliary
reconstruction module dedicated to pixel-wise refinement. The model
therefore prioritizes perceptually plausible and content-consistent
detail restoration, while retaining the decoded latent as the
structural anchor. Consequently, the main advantage of RIFT is more
pronounced in perceptual quality than in PSNR, which primarily
measures pointwise reconstruction error.

\section{Qualitative Comparison with Codec-only Reconstruction}
\label{app:rift_codec_only}

We compare RIFT with codec-only reconstruction to illustrate the
visual effect of generative restoration. Both use the same
codec and identical bitstreams. Codec-only directly decodes the
compressed latent, whereas RIFT decodes the restored latent.
The codec-only baseline therefore bypasses the DiT and all
restoration heads. No model is retrained for this comparison.

\begin{figure*}[p]
\centering
\includegraphics[width=\textwidth,height=0.82\textheight,keepaspectratio]
{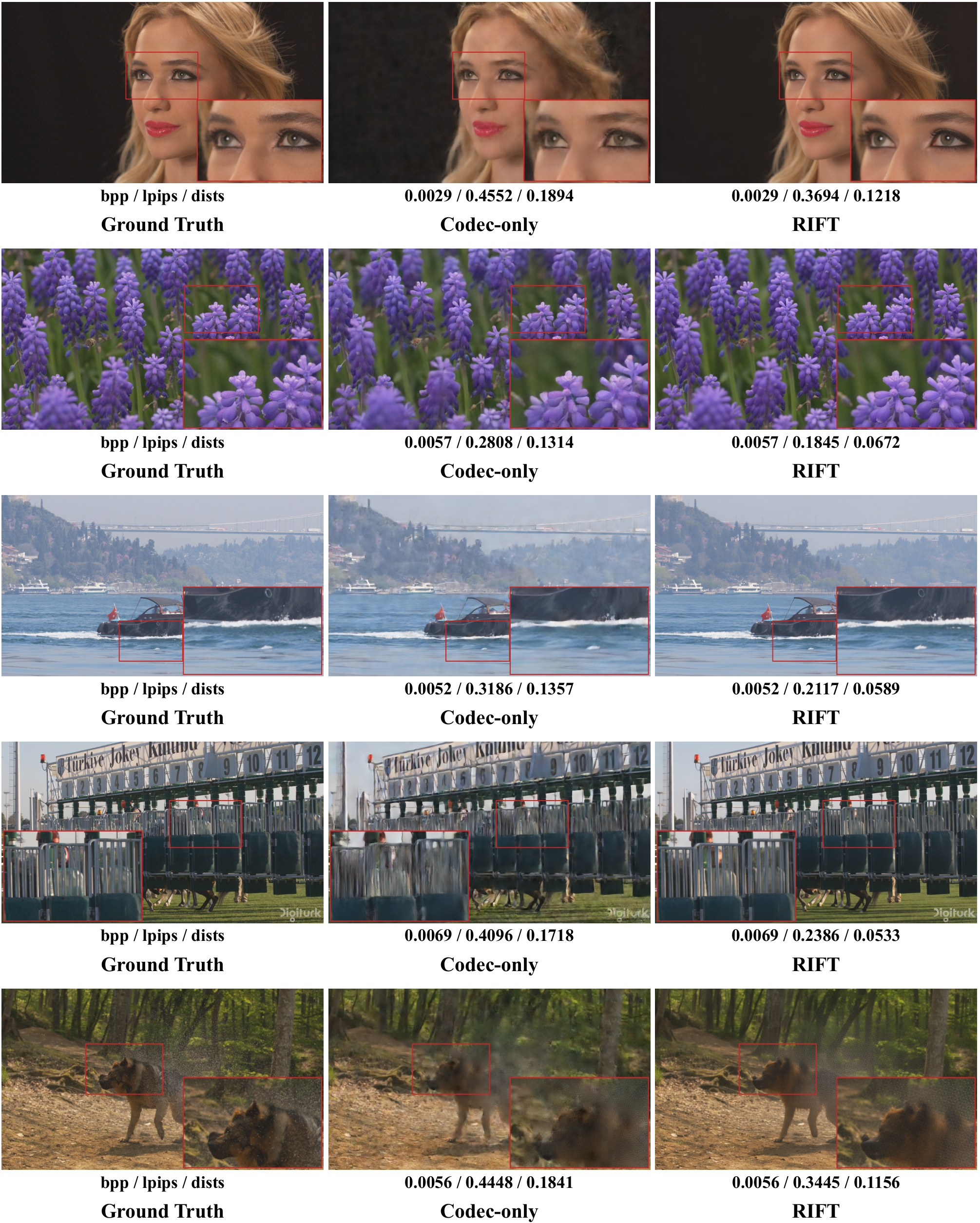}
\caption{Qualitative comparison with codec-only reconstruction at
QP 63. Rows show Beauty, HoneyBee, Bosphorus, ReadySteadyGo, and
ShakeNDry; columns show ground truth, codec-only reconstruction,
and RIFT. Red boxes mark the regions enlarged in the insets.
Codec-only and RIFT use identical bitstreams. Labels report actual
sequence bpp, full-frame LPIPS-VGG, and full-frame DISTS, respectively.
Bpp uses the coded $1088\times1920$ canvas; quality metrics are
computed on floating-point reconstructions cropped to
$1080\times1920$. Displayed images are exported as lossless frames.}
\label{fig:rift_codec_only_subjective}
\end{figure*}

Figure~\ref{fig:rift_codec_only_subjective} presents five UVG
sequences at 1080p: Beauty, HoneyBee, Bosphorus, ReadySteadyGo,
and ShakeNDry. We use the lowest measured rate setting (QP 63)
and show frame 48 (zero-based) from each sequence. The first 96
frames are coded with GOP lengths of 25, 25, 25, and 21.
Within each row, all methods show the same frame and spatial crop.
The enlarged regions illustrate how generative restoration improves
local structures and fine textures over direct codec-only
reconstruction in these examples, at identical bitrate.

\section{Limitations and Future Work}
\label{app:limitations}

RIFT demonstrates that codec-residual restoration can be effectively
guided by explicitly modulating the direction and magnitude of the
generative update. However, the current model primarily validates this
idea rather than defining its performance limit. The modulation network
contains only $1.69\,\mathrm{M}$ parameters, a small fraction of the Video
DiT backbone. More advanced architectures and training strategies may
allow the DiT itself to learn this residual-aware adaptation implicitly.

Several limitations remain. First, the current mapping from operating
points to flow timesteps is fixed and may not be optimal for different
video content or degradation patterns. Second, each GOP is processed
independently, without transferring temporal information across GOP
boundaries, which may limit long-term temporal consistency. Future work
will therefore explore adaptive timestep selection, implicit
residual-aware adaptation within the generative backbone, and cross-GOP
temporal modeling. We also plan to distill and optimize the VAE to reduce
its computational cost and further improve end-to-end reconstruction
speed.

\end{document}